\documentclass{article}

\usepackage{arxiv}

\usepackage[utf8]{inputenc} 
\usepackage[T1]{fontenc}    
\usepackage{hyperref}       
\usepackage{url}            
\usepackage{booktabs}       
\usepackage{amsfonts}       
\usepackage{nicefrac}       
\usepackage{microtype}      
\usepackage{lipsum}		
\usepackage{graphicx}
\usepackage{natbib}
\usepackage{doi}
\usepackage{pgfplots}
\pgfplotsset{compat=1.7}

\title{Succinct Differentiation of Disparate Boosting Ensemble Learning Methods for Prognostication of Polycystic Ovary Syndrome Diagnosis}

\author{Abhishek Gupta \\
	Department of EXTC \\
	University of Mumbai \\
	Mumbai, 400032, India \\
	\texttt{abhishekgupta@sjcem.edu.in} \\
	\And
	Sannidhi Shetty \\
	Department of Computer Engineering \\
	University of Mumbai \\
	Mumbai, 400032, India \\
	\texttt{shettysannu11@gmail.com} \\
	\And
	Raunak Joshi \\
	Department of Information Technology \\
	University of Mumbai \\
	Mumbai, 400032, India \\
	\texttt{raunakjoshi.m@gmail.com} \\
	\And
	Ronald Melwin Laban \\
	Department of EXTC \\
	St. John College of Engineering and Management \\
	Palghar, 401404, India \\
	\texttt{ronaldlaban@gmail.com} \\
}

\renewcommand{\shorttitle}{}

\begin{document}
\maketitle

\begin{abstract}
	Prognostication of medical problems using the clinical data by leveraging the Machine Learning
    techniques with stellar precision is one of the most important real world challenges at the present time.
    Considering the medical problem of Polycystic Ovary Syndrome also known as PCOS is an emerging
    problem in women aged from 15 to 49. Diagnosing this disorder by using various Boosting Ensemble
    Methods is something we have presented in this paper. A detailed and compendious differentiation
    between Adaptive Boost, Gradient Boosting Machine, XGBoost and CatBoost with their respective
    performance metrics highlighting the hidden anomalies in the data and its effects on the result is
    something we have presented in this paper. Metrics like Confusion Matrix, Precision, Recall, F1
    Score, FPR, RoC Curve and AUC have been used in this paper.
\end{abstract}


\section{Introduction}
The most common gynecological disorder affecting women globally is known as polycystic ovary syndrome (PCOS).
The symptoms of PCOS include irregular periods, hirsutism, thinning hair and hair loss over head, oily skin or acne
and weight gain. PCOS can lead to risk in later life with a lifelong situation that causes a person’s blood sugar levels
to promote type-II diabetes. High blood pressure and high cholesterol which can lead to heart stroke, overweight
ladies may expand sleep apnoea, a situation that causes interrupted breathing at some stage in sleep. Around 10 -
15\% of reproductive age (15 to 49 years) of women suffer from this. The monetary expenses of this disease and its
comorbidities need the development of instruments and techniques so one can permit for early and precise identification.
To cope with this problem this paper proposes a system for the early detection and prediction of PCOS from the most
reliable and minimal and promising scientific and metabolic parameters, which is early detection for these diseases.
Machine Learning\citep{8697857} can be leveraged to perform prognostication of PCOS that exigently extracts factual records from
the given statistics considering the fact that machine learning is better known as glorified statistics. A specific type of
machine learning algorithm that seeks to improve the overall performance by combining the predictions from more than
one model which is a trendy meta method is known as an Ensemble Learning Approach. The Ensemble Methods\citep{opitz1999popular}
work with analogy of working with multiple weak learners. These weak learners are a set of very negligible accuracy yielding classifiers. These weak learners later are combined to give the best possible result and turn them into strong
learners. The two main dominant classes of Ensemble Learning are Bagging\citep{Breiman2004BaggingP} and Boosting\citep{Freund1999ASI}.
Bagging Ensemble Method is also known as Bootstrap Aggregation Method. It works with the approach of accumulating
all the sets of weak learners and aggregating them later to give the best result. Random Forest\citep{breiman2001random} is the most common
type of Bagging Ensemble Model. Bagging easily tackles the problem of Bias and Variance Trade-off. Usually the
aggregation process of the Bagging is some sort of Deterministic Averaging method. Process of Bagging itself is the
averaging method where the algorithm fits all the weak learners. For classification methods in Bagging the voting
methods help determine the best possible outcome in a single value representation. Random Forest is a commonly used
Bagging approach where it helps overcome the issues with Decision Trees. High depth in trees can cause high issues
with Variance. Shallow trees perform better for ensemble methods as the depth is less which maintains variance but
ends up in high bias but the trade-off is manageable. Although random forest ends up considering high depth trees and
giving a very considerably good result.

There are many leaks in Bagging to some extent. In sequential methods the combined weak learners are no longer
fitting independently from each other. The idea is to fit models iteratively such that the training of models at a given step depends on the models fitting the previous steps. Boosting is the approach used and ends up producing an ensemble
model that is in general less biased than the weak learners that compose it. Shallow Trees are commonly used over here.
Boosting methods work in the same fashion as bagging methods by building an association of multiple models that are
aggregated to obtain a strong learner that performs better than weak learners. However, bagging mainly aims at reducing
variance, boosting is a technique that works in fitting sequentially multiple weak learners in a very adaptive fashion.
Every learner in the sequence is fit, giving uttermost importance to observations in the data that were inappropriately
managed by the previous learners in the sequence. Intuitively, every new learner focuses its direction on the most
complex observations to fit, so that we achieve at the end of the process, a strong learner with lower bias.

Another major cause that inspires researchers to use low variance but high bias models as weak learners for boosting
is they are less computationally inexpensive to fit. As the cost to fit the varied learners cannot be done in parallel
fashion, it could become too expensive to fit several complex learners. Once the weak learners are chosen, researchers
are required to specify how learners will sequentially fit weak learners and how they are going to get aggregated.
Some of the commonly used Boosting Techniques are Ada-Boost\citep{Freund1999ASI} and Gradient Boosting\citep{Friedman2001GreedyFA}. Adaptive boosting modifies the weights attached to every single of the training observations whereas gradient boosting updates the value of these observations. After many years the Gradient Boosting Algorithm went under many modifications and some
more efficient variants of the algorithm were discovered. The commonly found replacement for Gradient Boosting is
XGBoost\citep{chen2016xgboost} which is an abbreviation for Extreme Gradient Boosting. It is highly optimized as compared to traditional
Gradient Boosting. Its main features are parallel processing, tree pruning, handling missing values in efficient fashion
and inbuilt L2 Regularization\citep{ng2004feature}.

One more such improved variation of Gradient Boosting Machine is Cat Boost\citep{prokhorenkova2018catboost}. The cat indicates Categorical
Boosting. Performance of the Cat Boost is very much better than other Gradient Boosting Algorithms. It handles
Categorical Values automatically without any requirement of explicit preprocessing. It performs very efficient hyperparameter tuning and manages the over-fitting issues. It contains parameters like the Number of Trees, Learning Rate,
Regularization, Tree Depth, Fold Size and many other parameters that can be tuned.

\section{Methodology}
\subsection{Approach}
An intricate problem like PCOS prognostication has many facets and needs to be addressed with utmost precaution.
The integral part of such a problem lies under the roof of appropriately sampled dataset. The derived sample from the
population should give an appropriate standard scaled mechanism that follows the central limit theorem and derives
flawless semantics. Second part is the prediction metrics. We present multiple confusion matrices of different
algorithms that orchestrate the number of False Positives and False Negatives. PCOS prognostication requires the least
number of False Positives possible. The problem lies beneath the fact that if the person suffers from PCOS, the system
is supposed to give an accurate metric leading to further steps for treatment.

\subsection{Features}
The data to be considered for PCOS prognostication should consider some major features. These features extensively
will result in best prediction practices. Some of the features are Age, hair growth on different areas of the body based
on the intensity of disorder, lifestyle with relation to the activity levels, habit of ingesting and craving high carb junk
foods, weight boundaries, sudden weight gain symptoms, irregularities in menstrual cycle, conceiving status, acne or skin tagging, thinning of hair, dark patches over skin, fatigue or tiredness, mood swings. These are some of the most
common features taken into consideration for diagnosis of PCOS by medical professionals. The same features are
supposed to be given by the dataset we will use for prognostication.

\subsection{Problem Specification}
Choosing the right algorithm is the prime task of this problem statement. The specification of the problem can be done
as we are addressing binary classification problem. The reason for using binary classification specification is the
result which states whether the user has PCOS or not. Since there is an involvement of only discrete values affecting the
decision with a boundary of the probabilities lying between 0 to 1, states binary classification is the right choice. The
predicted probabilities are separated using a threshold value that creates a difference for final predictions which are
discrete values.

\section{Implementation}
The implementation section takes into consideration various types of Machine Learning state of the art algorithms. The
binary classification specification is taken into account while selecting all the Machine Learning algorithms.

\subsection{AdaBoost}
AdaBoost\citep{Freund1999ASI} is one of the preliminary Boosting Algorithms used in Ensemble Methods. It works with the indulgence of a
set of random forests in an association. This technique uses a random forest of stumps. These stumps are single node
trees that can help determine the decision on a very nominal level. These stumps are weak classifiers and when their
result is added together turn out to be a strong classifier. The output of the hypothesis produced by these weak classifiers
are represented by $h(xi)$ for each sample in the training set. A coefficient is assigned to a selected weak classifier for
each iteration t such that, the sum of the training error $E$ of the resulting t-stage boost classifier is minimized. We also used the random state parameter of the stumps to 42 which was able to yield us the best possible result.

\subsection{Gradient Boosting}
Gradient Boosting\citep{Friedman2001GreedyFA} is preferred in many cases over AdaBoost due to its Greedy Algorithmic approach. But this can be
detrimental in some scenarios. Regularization Practice can although make it a quite conventional Boosting Algorithm.
It calculates residuals which are modulated in a prescribed fashion. Inflection Point is taken into consideration to avoid
over-fitting. We went with the learning rate as 0.01 which was able to yield a very good result.

\subsection{XGBoost}
XGBoost\citep{chen2016xgboost} uses the basic structure of Gradient Boosting Algorithm. It is highly optimized and solves various issues
recurring with the Gradient Boost. It performs Regularization by default. It handles the missing data values quite
efficiently. It is very subtle to cross validation. Its main feature is Parallelized Tree Association and Tree Pruning for
depth management. It also has very good cache awareness. Even with less hyper-parameter tuning it was able to provide
very good results.

\subsection{CatBoost}
After the successful development of the XGBoost many more modifications were brought into the arsenal of Boosting
Algorithms. CatBoost\citep{prokhorenkova2018catboost} is associated with the categorical variables on a very high scale. It provides you the leverage of
giving indices of categorical columns. It one hot encodes for the number of features with a similar number of values
less than or equal to their given parameter values. This encoding capability easily overcomes the over-fitting problem. It
is similar to the mean encoding. It used L2 Regularization\citep{ng2004feature}. The specified depth value is 16 but recommended is
1–10. The features are used on the basis of each split. This is calculated using the Random Subspace Method. The
training time taken by CatBoost is quite less which is definitely an added benefit.

\subsection{Explanatory Variables}
These are the set of features we have taken into consideration for the training purpose. The features used represent
the typical parameters to determine if the patient has PCOS. The prognostication expects features to highlight the
best possible factors in detection. The most common features used are Age, Weight condition, unexpected sudden hair growth in various areas of the body, lifestyle beyond or less than sedentary, irregularities in periods or have been
conceived before.

\section{Results}
This is the section of the paper that holds the entire essence of our paper. This gives a deep insight about the differences in all of the boosting algorithms we have taken into consideration. The results solely lie on the basis of the performance metrics on the validation data.

\subsection{Train and Test Accuracy}
This is one of the preliminary metrics used as a result to hypothesize the current goal. The training and testing accuracy gives a generic measure of achievable result by the algorithm. It not only specifies the current state of the algorithm but also sets the boundary for further development in the algorithm.

\begin{table}[htbp]
    \centering
    \caption{Comparison of Various Boosting Techniques Used}
    \begin{tabular}{lll}
        \toprule
        Algorithm & Training & Testing \\
        \midrule
        AdaBoost & 96.49\% & 85.71\% \\
        Gradient Boosting & 88.6\% & 79.59\% \\
        XGBoost & 95.61\% & 85.71\% \\
        \textbf{CatBoost} & \textbf{99.98\%} & \textbf{95.92\%} \\
        \bottomrule
    \end{tabular}
    \label{tab:my_label}
\end{table}

The table \ref{tab:my_label} gives the set of accuracies and we can clearly see CatBoost has very less variation between the training and testing accuracy making it a preferable algorithm for the specified problem. But this certainly is not a specified approach as accuracy metric alone cannot give the right consideration when it is about the compendious comparison of various similarly competitive algorithms.

\subsection{Confusion Matrix}
In this paper we are trying to aim for a problem that is prognostication of PCOS and it requires a very heavy emphasis on a particular set of errors. Statistically there are 2 types of errors which require to be targeted with respect to the problem. These errors are Type-I Error and Type-II Error. These errors are unavoidable in any problem and given consideration to our problem, correct detection of the PCOS is essential.

\begin{table}[htbp]
    \centering
    \caption{Confusion Matrix}
    \begin{tabular}{lll}
        \toprule
        Algorithm & False Negatives &  False Positives \\
        \midrule
        AdaBoost & 3 & 8 \\
        Gradient Boosting & 7 & 2 \\
        XGBoost & 5 & 4 \\
        \textbf{CatBoost} & \textbf{1} & \textbf{1} \\
        \bottomrule
    \end{tabular}
    \label{tab:my_label2}
\end{table}

Assuming the problem is that the patient suffers from PCOS the system is supposed to correctly identify the problem for further measures. This category falls clearly under the Type-II Error. This error is also specifically targeting the False Negatives. This is where the Confusion Matrix\citep{Ting2010ConfusionM} comes into the picture as the ratio of the values of both types of errors with their respective true states. Clearly the above table proves that CatBoost is the most preferable algorithm and the accuracy is not only an important measure which we can use as our basis. The total values taken into the test set are 48 records where CatBoost performs the best with the least number of False Negatives which is something we are specifically targeting.

\subsection{Metric Scores}
Some majorly used metrics in determining the better performance of machine learning algorithms are never complete without Precision, Recall and F-Score\citep{powers2020evaluation}. Precision is a measure of quality of predictions. It is denoted by formula as

\begin{equation}
    Precision = \frac{TP}{TP + FP}
\end{equation}

Recall is a measure of the quantity of correctly classified predictions. It is denoted by formula as

\begin{equation}
    Recall = \frac{TP}{TP + FN}
\end{equation}

F-Score gives a statistical measure of test-set for binary classification values. It is denoted by formula as

\begin{equation}
    F-Score = 2*\frac{P*R}{P+R}
\end{equation}

It is very important because it can maintain the balance between Precision and Recall. It works best for uneven data
distribution

\begin{table}[htbp]
    \centering
    \caption{Confusion Matrix}
    \begin{tabular}{llll}
        \toprule
        Algorithm & Precision & Recall & F-Score \\
        \midrule
        AdaBoost & 0.9069 & 0.975 & 0.9397 \\
        Gradient Boosting & 0.8695 & 0.944 & 0.9302 \\
        XGBoost & 0.9090 & 0.944 & 0.9523 \\
        \textbf{CatBoost} & \textbf{0.9523} & \textbf{1.0} & \textbf{0.9756} \\
        \bottomrule
    \end{tabular}
    \label{tab:my_label3}
\end{table}

We have also given some results of precision recall and their trade-off. Precision Recall Curve gives detailed flow of the values. Precision Recall Curve is used mostly when there is large imbalance in dataset and we want to practically harness our problem to find which is the best choice for Boosting Algorithm since relying on a single metric is definitely detrimental and not a proper measure of selecting the algorithm.

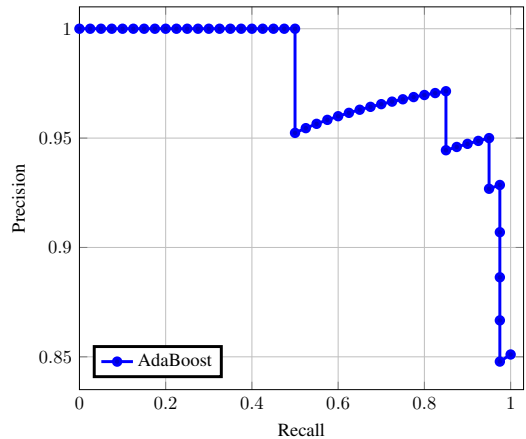
\begin{figure}[htbp]
    \centering
    \begin{tikzpicture}[scale=0.7]
        \begin{axis}[xlabel={Recall},ylabel={Precision},enlargelimits=false,
        grid=both, ymax=1.01, ymin=0.835, xmax=1.03,
        scale only axis=true,legend pos=south west,style={ultra thick}, axis line style={ultra thin}]
        \addplot+[blue] table[x=ada_recall,y=ada_precision,col sep=comma]{plots/pr/adapr.csv};
        \addlegendentry{AdaBoost}
        \end{axis}
    \end{tikzpicture}
    \caption{Precision Recall Curve for AdaBoost}
    \label{fig:a}
\end{figure}

In figure \ref{fig:a} the Precision and Recall Curve is quite varied. A significant drop of the Precision Recall trade-off can also be seen. In figure \ref{fig:b} the distance between Precision and Recall varies significantly. Fluctuating accuracies show that large imbalance in the data can affect the GBM on a greater scale. The values even tried rising from the steep position but ultimately reached the lowest point.

\begin{figure}[htbp]
    \centering
    \begin{tikzpicture}[scale=0.7]
        \begin{axis}[xlabel={Recall},ylabel={Precision},enlargelimits=false,
        grid=both, ymax=1.005, ymin=0.94, xmax=1.03,
        scale only axis=true,legend pos=south west,style={ultra thick}, axis line style={ultra thin}]
        \addplot+[red] table[x=gbm_recall,y=gbm_precision,col sep=comma]{plots/pr/gbmpr.csv};
        \addlegendentry{Gradient Boosting}
        \end{axis}
    \end{tikzpicture}
    \caption{Precision Recall Curve for GBM}
    \label{fig:b}
\end{figure}
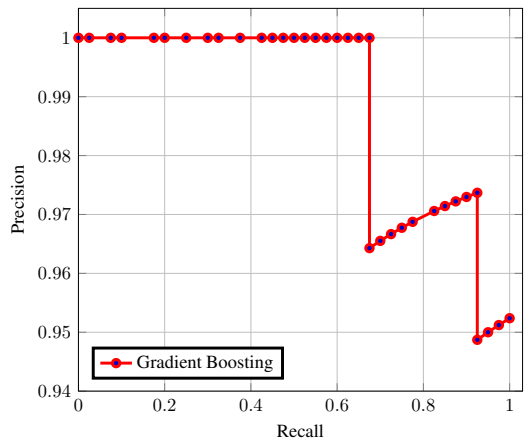

\begin{figure}[htbp]
    \centering
    \begin{tikzpicture}[scale=0.7]
        \begin{axis}[xlabel={Recall},ylabel={Precision},enlargelimits=false,
        grid=both, ymax=1.009, ymin=0.91, xmax=1.03,
        scale only axis=true,legend pos=south west,style={ultra thick}, axis line style={ultra thin}]
        \addplot+[brown] table[x=xg_recall,y=xg_precision,col sep=comma]{plots/pr/xgpr.csv};
        \addlegendentry{XGBoost}
        \end{axis}
    \end{tikzpicture}
    \caption{Precision Recall Curve for XGBoost}
    \label{fig:c}
\end{figure}
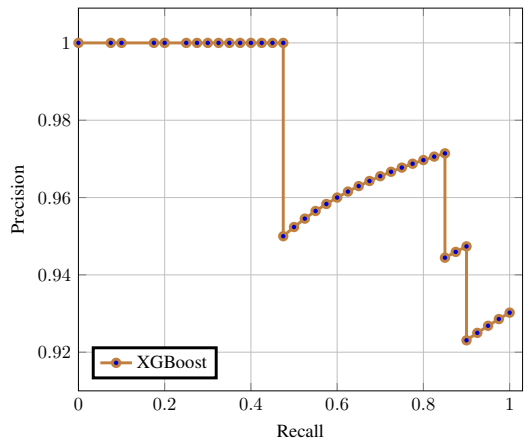

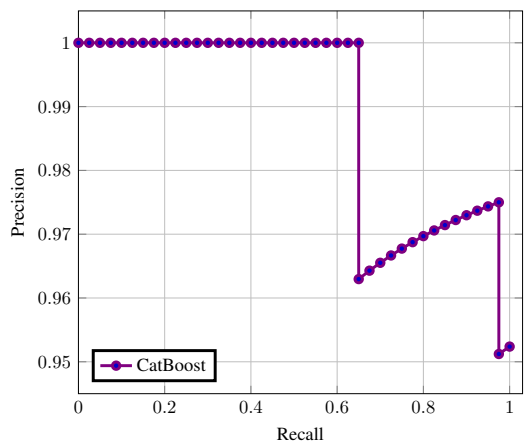
\begin{figure}[htbp]
    \centering
    \begin{tikzpicture}[scale=0.7]
        \begin{axis}[xlabel={Recall},ylabel={Precision},enlargelimits=false,
        grid=both, ymax=1.005, ymin=0.945, xmax=1.03,
        scale only axis=true,legend pos=south west,style={ultra thick}, axis line style={ultra thin}]
        \addplot+[violet] table[x=cat_recall,y=cat_precision,col sep=comma]{plots/pr/catpr.csv};
        \addlegendentry{CatBoost}
        \end{axis}
    \end{tikzpicture}
    \caption{Precision Recall Curve for Cat Boost}
    \label{fig:d}
\end{figure}

In figure \ref{fig:c} the distance between Precision and Recall varies but it may seem better than the GBM.
Fluctuating accuracies show that large imbalance does not affect the XGBoost as much as the GBM Classifier. The
accuracies are obviously varied but are better than AdaBoost. It is not the number that matters but the difference which can maintain the trade-off. In figure \ref{fig:d} the distance between Precision and Recall varies the least. This indicates that it is considerably better than other Boosting mechanisms taken into account. The CatBoost not only maintains the least difference but also shows varied Precision Recall Curve trade-off in an efficient manner. It closes major gaps in classification continuously making it a better choice for the given problem.

\subsection{RoC and AUC}
RoC Curve\citep{Fawcett2006AnIT} is known as Receiver Operator Characteristic Curve which is used for performance measurement for
classification models. It gives detailed probability fluctuations throughout the predictions indicating the classes. This is calculated on the basis of 2 pillars known as True Positive Rate (TPR) and False Positive Rate (FPR). TPR is also
known as Sensitivity and works much similar to Recall. It is denoted by formula as

\begin{equation}
    TPR = \frac{TP}{TP + FN}
\end{equation}

After TPR we have to work our way towards False Positive Rate, but the problem is it cannot be calculated directly. We need to first find Specificity. It is a measure for finding negative classes without anomalies. This is represented by formula as

\begin{equation}
    Specificity = \frac{TN}{TN + FP}
\end{equation}

Now this specificity can help us find the False Positive Rate. All we need to do is subtract the Specificity from 1. This gives us FPR but it can be also denoted with a formula

\begin{equation}
    FPR = \frac{FP}{FP + TN}
\end{equation}

This gives us the RoC Curve but that is not enough for speculating the algorithm’s performance. We need AUC which is also known as Area Under Curve\citep{Bradley1997TheUO}. This AUC helps specify threshold settings. It also represents the degree or measure of separability. Higher the value of AUC, better is the prediction. We require our algorithm to bypass the value set on the prediction scale from 0 to 1. The highest value is considered to be the best algorithmic performance. RoC Curve and AUC are determined to give the best performance metric among other methods.

\begin{figure}[htbp]
    \centering
    \begin{tikzpicture}[scale=0.8]
        \begin{axis}[xlabel={False Positive Rate},ylabel={True Positive Rate},enlargelimits=false,
        grid=both, ymax=1.05, xmax=1.03, xmin=-0.05, ymin=-0.05,
        scale only axis=true,legend pos=south east,style={ultra thick}, axis line style={ultra thin}]
        \addplot+[blue] table[x=fpr,y=tpr,col sep=comma]{plots/roc/adaroc.csv}; 
        \addplot+[red] table[x=fpr,y=tpr,col sep=comma]{plots/roc/gbroc.csv}; 
        \addplot+[brown] table[x=fpr,y=tpr,col sep=comma]{plots/roc/xgbroc.csv};
        \addplot+[violet] table[x=fpr,y=tpr,col sep=comma]{plots/roc/catroc.csv};
        \addplot+[black] coordinates { (0,0) (1,1) };
        \addlegendentry{AdaBoost}
        \addlegendentry{Gradient Boosting}
        \addlegendentry{XGBoost}
        \addlegendentry{CatBoost}
        \end{axis}
    \end{tikzpicture}
    \caption{RoC and AUC}
    \label{fig:e}
\end{figure}
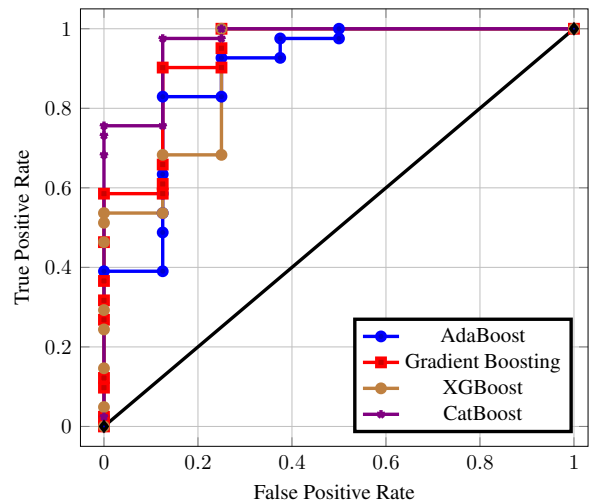

In figure \ref{fig:e} the black solid line indicates the threshold boundary for the AUC. All the curves need to be above the threshold and so has happened. The AUC gives detailed fluctuations of all the probabilities at every instance and CatBoost proves to be the best performer definitely. The area covered is the highest for the CatBoost. XGBoost in the beginning covers the least area but later bypasses the Gradient Boost. AdaBoost is considered to be the second best choice for the Area covered. In the beginning it went along with the CatBoost but later fell short yet managed to cover a greater amount of Area as compared to the Gradient Boosting and XGBoost making it the second preferred algorithm.

\section{Conclusion}
This paper outlines the compendious comparison of contrasting Boosting Ensemble Methods in detail with performance
metrics for prognostication of PCOS Diagnosis. Considering the Boosting Algorithms we were able to present the best
of most commonly used algorithms, viz. AdaBoost, Gradient Boosting Machine, Extreme Gradient Boosting and Cat Boost. Our result section was the primary essence of this paper, because it was able to give a detailed explanation with
respect to multiple facets for assessing an algorithm, viz. Precision, Recall, F1-Score, Confusion Matrix, False Positive
Rate, True Positive Rate and RoC Curve with AUC. Definitely this is not where the improvement in the paper stops,
but our work has opened a new gateway for emerging ideas by enthusiasts. With our best belief and knowledge we
conclude this paper.

\bibliographystyle{unsrtnat}
\bibliography{references}  






\end{document}